\documentclass[11pt]{article}
\usepackage{algorithm}
\usepackage{amsmath}
\usepackage{algpseudocode}
\usepackage{natbib}

\usepackage[final]{acl}

\usepackage{times}
\usepackage{latexsym}

\usepackage[T1]{fontenc}

\usepackage[utf8]{inputenc}

\usepackage{microtype}

\usepackage{inconsolata}

\usepackage{graphicx}

\usepackage{booktabs}
\usepackage{multirow}

\title{Not All Pretraining are Created Equal: Threshold Tuning and Class Weighting for Imbalanced Polarization Tasks in Low-Resource Settings}

\author{Abass Oguntade \\
    African Institute of Mathematical Sciences, South Africa \\
  \texttt{abasso@aims.ac.za} \\}

\begin{document}
\maketitle

\begin{abstract}
This paper describes my submission to the Polarization Shared Task at SemEval-2025, which addresses polarization detection and classification in social media text. I develop Transformer-based systems for English and Swahili across three subtasks: binary polarization detection, multi-label target type classification, and multi-label manifestation identification. The approach leverages multilingual and African language-specialized models (mDeBERTa-v3-base, SwahBERT, AfriBERTa-large), class-weighted loss functions, iterative stratified data splitting, and per-label threshold tuning to handle severe class imbalance. The best configuration, mDeBERTa-v3-base, achieves 0.8032 macro-F1 on validation for binary detection, with competitive performance on multi-label tasks (up to 0.556 macro-F1). Error analysis reveals persistent challenges with implicit polarization, code-switching, and distinguishing heated political discourse from genuine polarization.
\end{abstract}

\section{Introduction}

Attitude polarization, the sharp division of opinions into opposing groups characterized by hostility toward out-groups and blind in-group support, has become a defining feature of contemporary social media discourse . This phenomenon manifests in contexts such as elections, conflicts, protests, and political debates, where content may stereotype, vilify, dehumanize, or invalidate individuals based on their identities or affiliations. Detecting and understanding polarized content across linguistic and cultural contexts is critical for mitigating online harm and fostering constructive dialogue \cite{grimminger-klinger-2021-hate,ousidhoum-etal-2019-multilingual}. This shared task evaluates systems on polarization detection in English and Swahili social media text across three progressively specialized subtasks: (1) binary classification of polarized versus non-polarized content, (2) multi-label identification of polarization target types (political, racial/ethnic, religious, gender/sexual, other), and (3) multi-label detection of polarization manifestations (stereotype, vilification, dehumanization, extreme language, lack of empathy, invalidation) .

My main strategy combines systematic evaluation of six Transformer architectures: three multilingual models (TwHIN-BERT \cite{zhang-etal-2022-twhin}, DistilBERT-multilingual \cite{sanh2019distilbert}, mDeBERTa-v3-base \cite{he2021deberta}) and three African language-specialized models (SwahBERT \cite{martin-etal-2022-swahbert}, AfriBERTa-large \cite{ogueji-etal-2021-small}, AfroXLMR-large \cite{alabi-etal-2022-adapting}) with class-weighted loss functions and per-label threshold optimization for multi-label tasks. I employ iterative stratified splitting \cite{sechidis2011stratification,szymanski2017network} to preserve multi-label distributions, BCEWithLogitsLoss with per-label positive class weights to emphasize rare categories, and two-stage threshold search (coarse global + fine per-label) \cite{saito2015precision} to maximize macro-F1 across severely imbalanced labels.

The key findings reveal that architectural selection matters more than language-specific pretraining: mDeBERTa-v3-base outperforms Swahili-specialized models by 10--15 percentage points despite having no explicit Swahili focus. Threshold tuning provides substantial gains (20+ points macro-F1) for multi-label classification under extreme imbalance \cite{buda2018systematic}. However, the system struggles with implicit polarization cues, code-switching between English and Swahili, and false positives on heated but non-polarized political rhetoric. On official test sets in \texttt{Codabench}\footnote{\url{https://www.codabench.org/}}, I achieve 0.815 macro-F1 (English) and 0.785 macro-F1 (Swahili) for binary detection, with final multi-label scores of 0.464 (English) and 0.556 (Swahili) for manifestation identification.

The code are available at {[https://github.com/HayBeeCoder/polarization-detection]}.

\section{Background}

The Polarization Shared Task defines polarization as content displaying negative attitudes toward out-groups while showing solidarity with in-groups, encompassing stereotyping, vilification, dehumanization, intolerance, and divisive rhetoric. The task provides labeled social media datasets for English (3,222 training instances) and Swahili (6,991 training instances) across three hierarchical subtasks.

\subsection{Task Description}

\textbf{Subtask 1: Binary Polarization Detection.} Given a social media post, predict whether it contains polarized content (label=1) or not (label=0). For example, the post \textit{``These people are destroying our country with their backwards beliefs''} would be labeled as polarized (1), while \textit{``I disagree with this policy''} would be non-polarized (0). English exhibits moderate class imbalance (64\% non-polarized, 36\% polarized, 1.78:1 ratio), while Swahili shows near-perfect balance (50\% each class).

\textbf{Subtask 2: Multi-Label Target Type Classification.} For polarized texts, predict which of five target categories apply: \textit{political}, \textit{racial/ethnic}, \textit{religious}, \textit{gender/sexual}, or \textit{other}. Multiple labels may apply simultaneously. For example, \textit{``Muslim immigrants are ruining European values''} targets both \textit{racial/ethnic} and \textit{religious} categories. The English dataset shows extreme imbalance: political dominates with 1,150 instances (30.3\%), while gender/sexual appears in only 72 instances (1.9\%).

\textbf{Subtask 3: Multi-Label Manifestation Identification.} Predict how polarization manifests across six categories: \textit{stereotype}, \textit{vilification}, \textit{dehumanization}, \textit{extreme\_language}, \textit{lack\_of\_empathy}, or \textit{invalidation}. For instance, \textit{``Those animals don't deserve to be treated like humans''} exhibits \textit{dehumanization}, \textit{vilification}, and \textit{extreme\_language}. Vilification dominates with 3,741 instances, while dehumanization appears least frequently (1,284 instances).

The organizers provide official train/dev splits and evaluate using macro-averaged F1 as the primary metric to weight all labels equally despite severe imbalance. I participate in all three subtasks for both English and Swahili.

\subsection{Related Work}

Hate speech and polarization detection have been extensively studied for high-resource languages \cite{fortuna-nunes-2018-survey,vidgen-derczynski-2020-directions}, but African languages remain underexplored \cite{caselli-etal-2021-hatebert}. Recent work on multilingual hate speech demonstrates that cross-lingual transfer from English to low-resource languages often underperforms language-specific models when sufficient target-language data exists \cite{bigoulaeva-etal-2021-cross}. African language-focused models like SwahBERT \cite{martin-etal-2022-swahbert} and AfriBERTa \cite{ogueji-etal-2021-small} aim to address this gap through targeted pretraining on 11 African languages including Swahili, Hausa, and Amharic. This work contributes systematic comparisons between multilingual and African language-specialized architectures on polarization detection, revealing counterintuitive results where general multilingual models outperform specialized alternatives.

\section{Data Analysis}

I conduct exploratory analysis to understand dataset characteristics and inform modeling decisions.

\subsection{Dataset Composition}

\begin{figure}[t]
\centering
\includegraphics[width=0.45\textwidth]{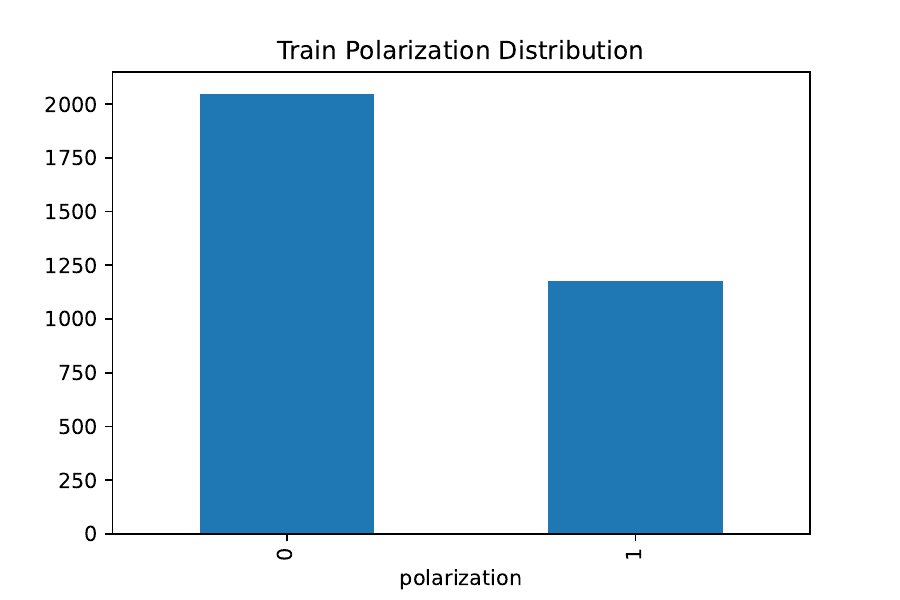}
\caption{English dataset polarization class distribution for Subtask 1. The dataset shows moderate imbalance with 64\% non-polarized instances.}
\label{fig:eng-class-dist}
\end{figure}

\begin{figure}[t]
\centering
\includegraphics[width=0.45\textwidth]{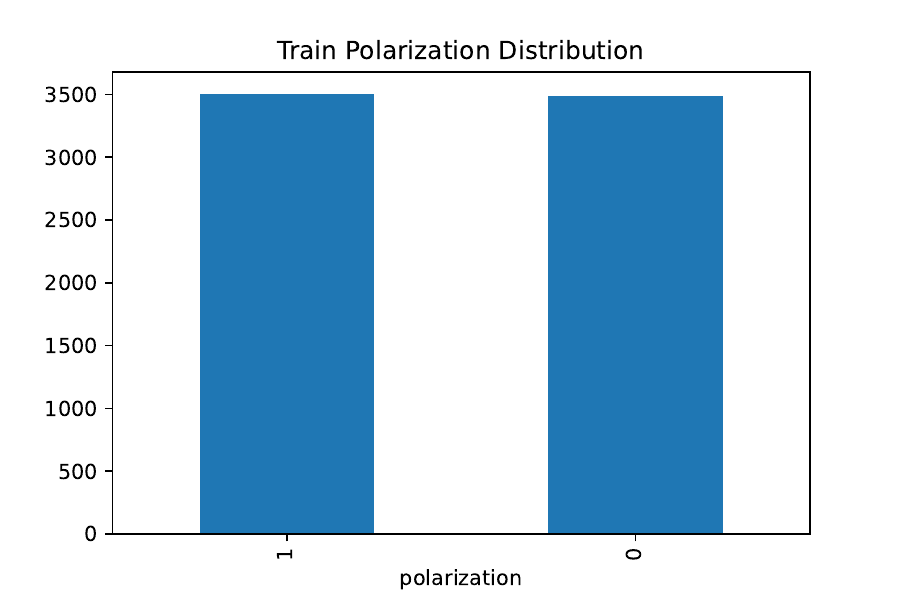}
\caption{Swahili dataset polarization class distribution for Subtask 1. Nearly balanced with 50\% representation in each class.}
\label{fig:swa-class-dist}
\end{figure}

Considering the separate English and Swahili as a whole, the complete dataset used comprises 10,213 instances: Swahili (6,991, 68.5\%) and English (3,222, 31.5\%). Figure~\ref{fig:eng-class-dist} and Figure~\ref{fig:swa-class-dist} show the class distributions for Subtask 1. English exhibits a 1.78:1 imbalance favoring non-polarized content, while Swahili shows a near-perfect balance.

\subsection{Vocabulary Patterns}

\begin{figure}[t]
\centering
\includegraphics[width=0.45\textwidth]{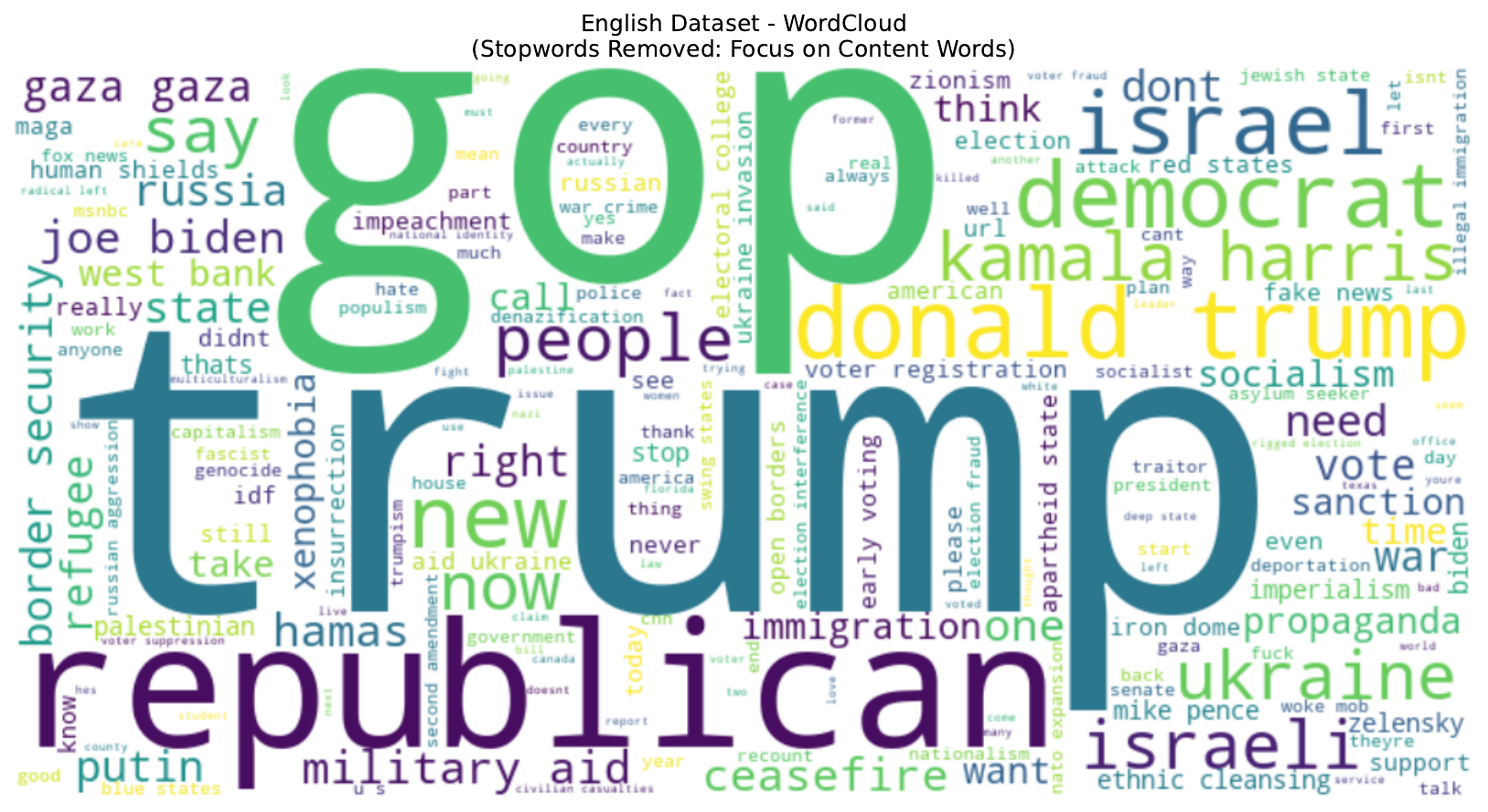}
\caption{Word cloud for English dataset showing dominance of political terms (trump, ukraine, gaza) and geopolitical discourse markers.}
\label{fig:eng-wordcloud}
\end{figure}

\begin{figure}[t]
\centering
\includegraphics[width=0.45\textwidth]{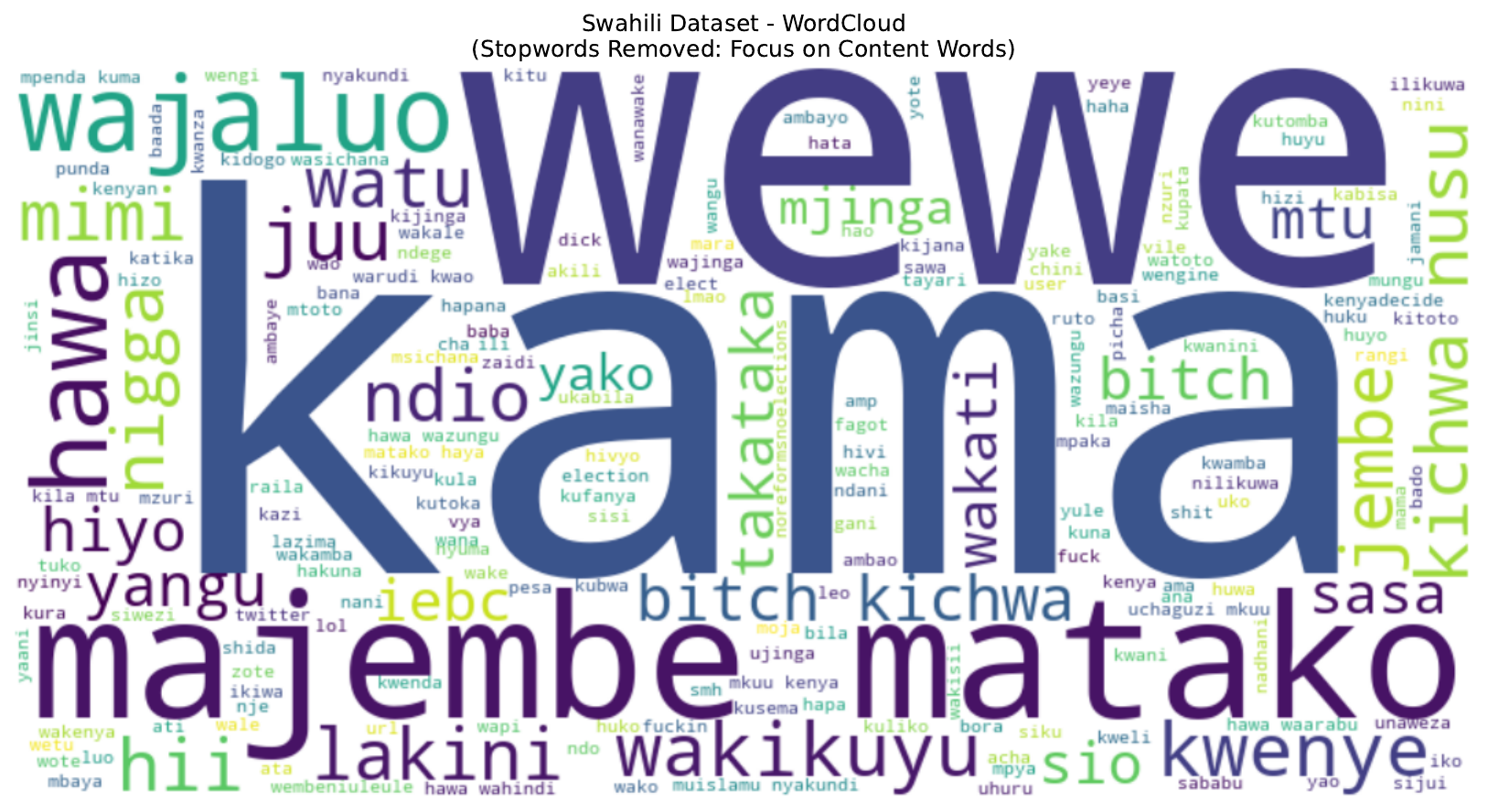}
\caption{Word cloud for Swahili dataset revealing frequent vulgar and offensive terms indicating different polarization expression patterns.}
\label{fig:swa-wordcloud}
\end{figure}

Figures~\ref{fig:eng-wordcloud} and~\ref{fig:swa-wordcloud} show word clouds revealing distinct topical focuses. Swahili content features vulgar terms (\textit{kutombwa}, \textit{hawa}, \textit{kama}, \textit{wewe}), while English centers on political discourse (\textit{trump}, \textit{ukraine}, \textit{gaza}, \textit{state}). This divergence suggests that cross-lingual transfer must account for culturally specific linguistic markers \cite{conneau-etal-2020-unsupervised}.

\subsection{Multi-Label Imbalance}

\begin{figure}[t]
\centering
\includegraphics[width=0.45\textwidth]{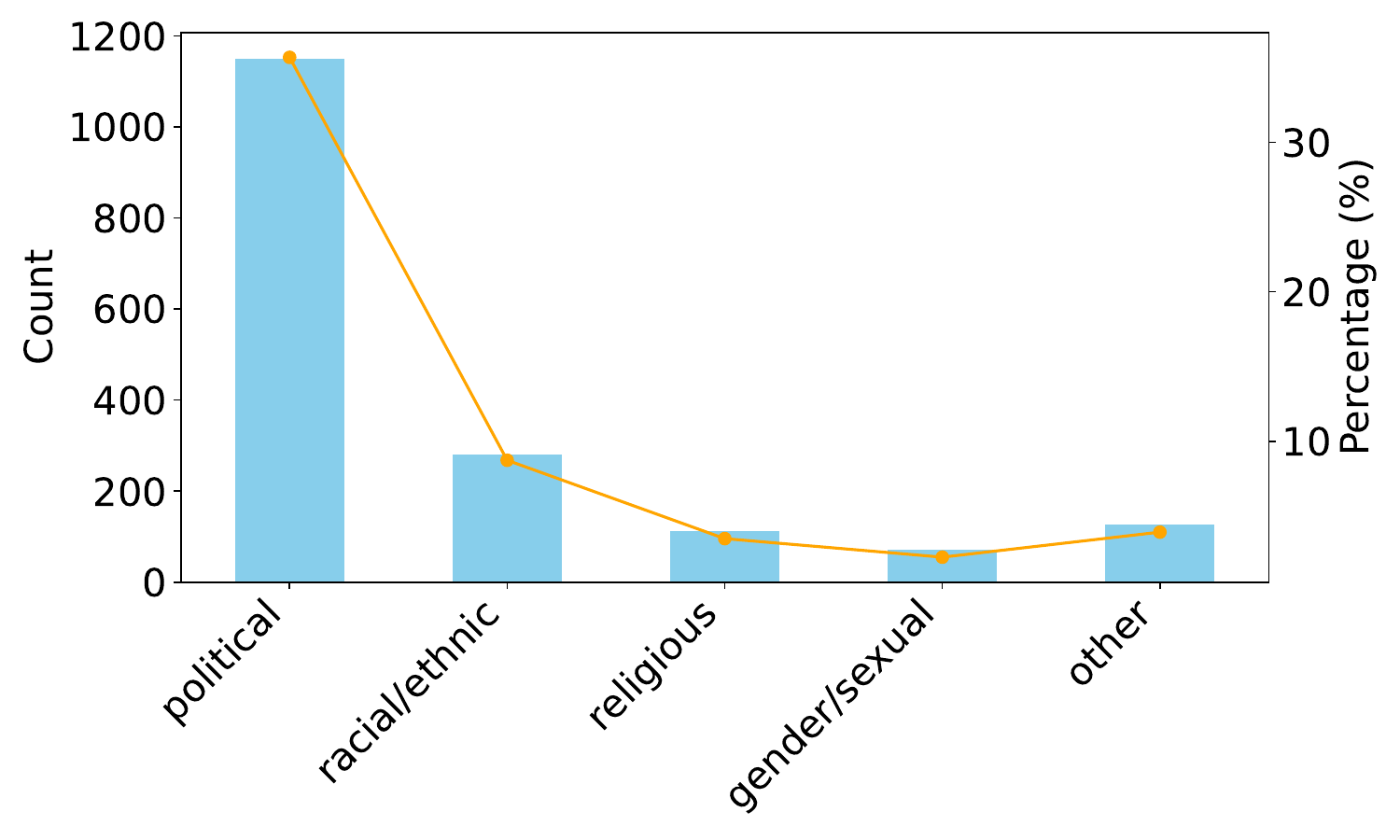}
\caption{Subtask 2 label distribution across five target types showing severe imbalance. Political dominates while gender/sexual remains extremely rare.}
\label{fig:label-dist-subtask2}
\end{figure}

Figure~\ref{fig:label-dist-subtask2} shows Subtask 2 label distribution for polarized texts in the English dataset. Political content dominates (35.7\% English), creating a 16.23:1 imbalance ratio with the rarest gender/sexual label (2.2\%). Notably, approximately 47\% of instances contain no active labels (i.e., all-zero class). This extreme imbalance motivates the use of class-weighted loss functions and threshold tuning strategies \cite{chawla2002smote,lin2017focal}.

\begin{figure}[t]
\centering
\includegraphics[width=0.45\textwidth]{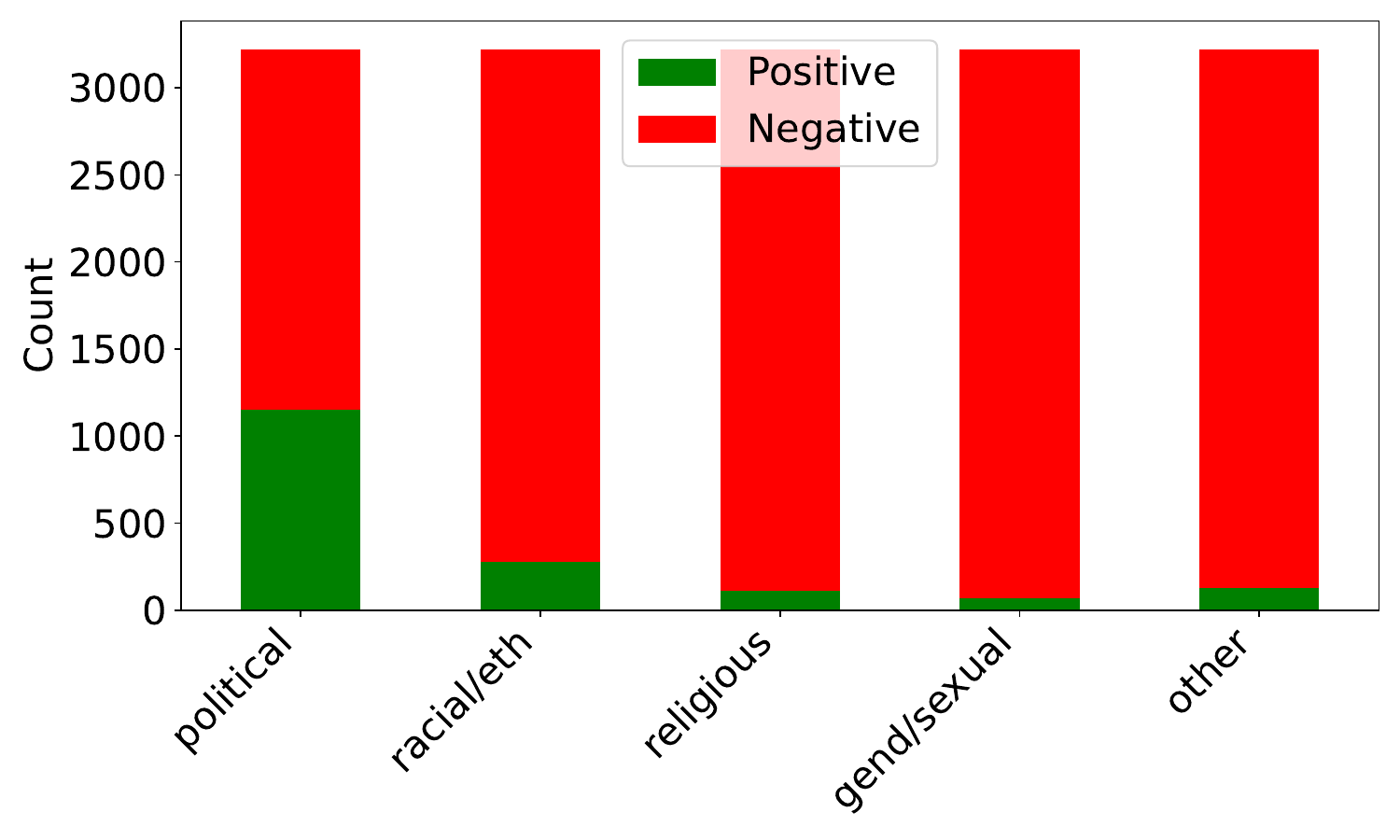}
\caption{Class distribution per label for Subtask 2 showing positive vs negative instances. All labels exhibit severe imbalance favoring negative class.}
\label{fig:class-balance-subtask2}
\end{figure}

Figure~\ref{fig:class-balance-subtask2} illustrates the class balance per label. Across all labels, negative (non-hate) examples vastly outnumber positive instances, with the 2,047 all-zero rows further amplifying skew toward non-hate content.

\subsection{Label Correlations}

\begin{figure}[t]
\centering
\includegraphics[width=0.45\textwidth]{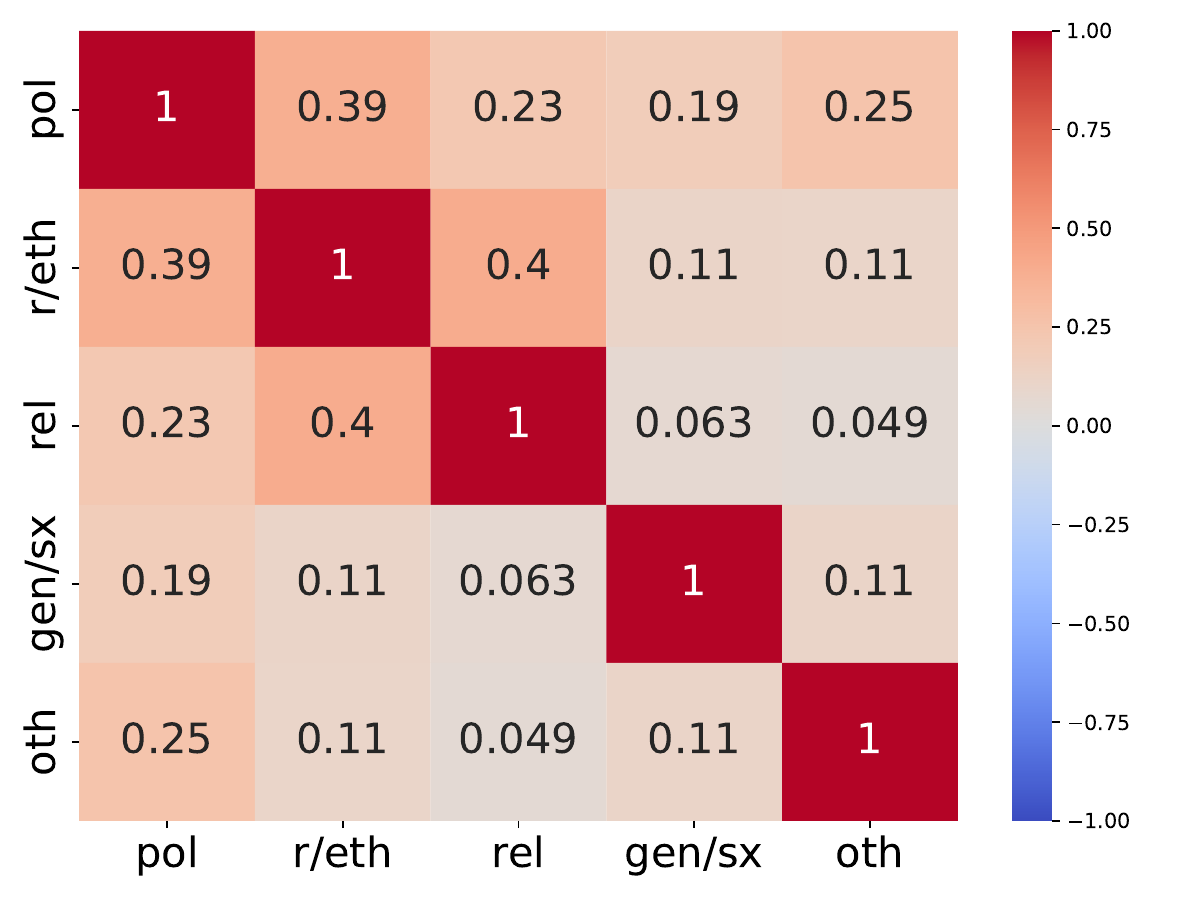}
\caption{Correlation heatmap for Subtask 2 labels. Weak correlations indicate mostly independent label relationships, except for racial/ethnic and religious (r=0.34).}
\label{fig:correlation-heatmap}
\end{figure}

Figure~\ref{fig:correlation-heatmap} shows correlation analysis revealing mostly near-zero correlations between label pairs in the English dataset. The notable exception is a moderate positive correlation (r=0.4) between racial/ethnic (r/eth) and religious (rel) labels, suggesting these categories co-occur in messages targeting ethnic-religious intersections. This finding aligns with prior work on intersectional hate speech \cite{ousidhoum-etal-2019-multilingual}.

\subsection{Manifestation Distribution}

\begin{figure}[t]
\centering
\includegraphics[width=0.45\textwidth]{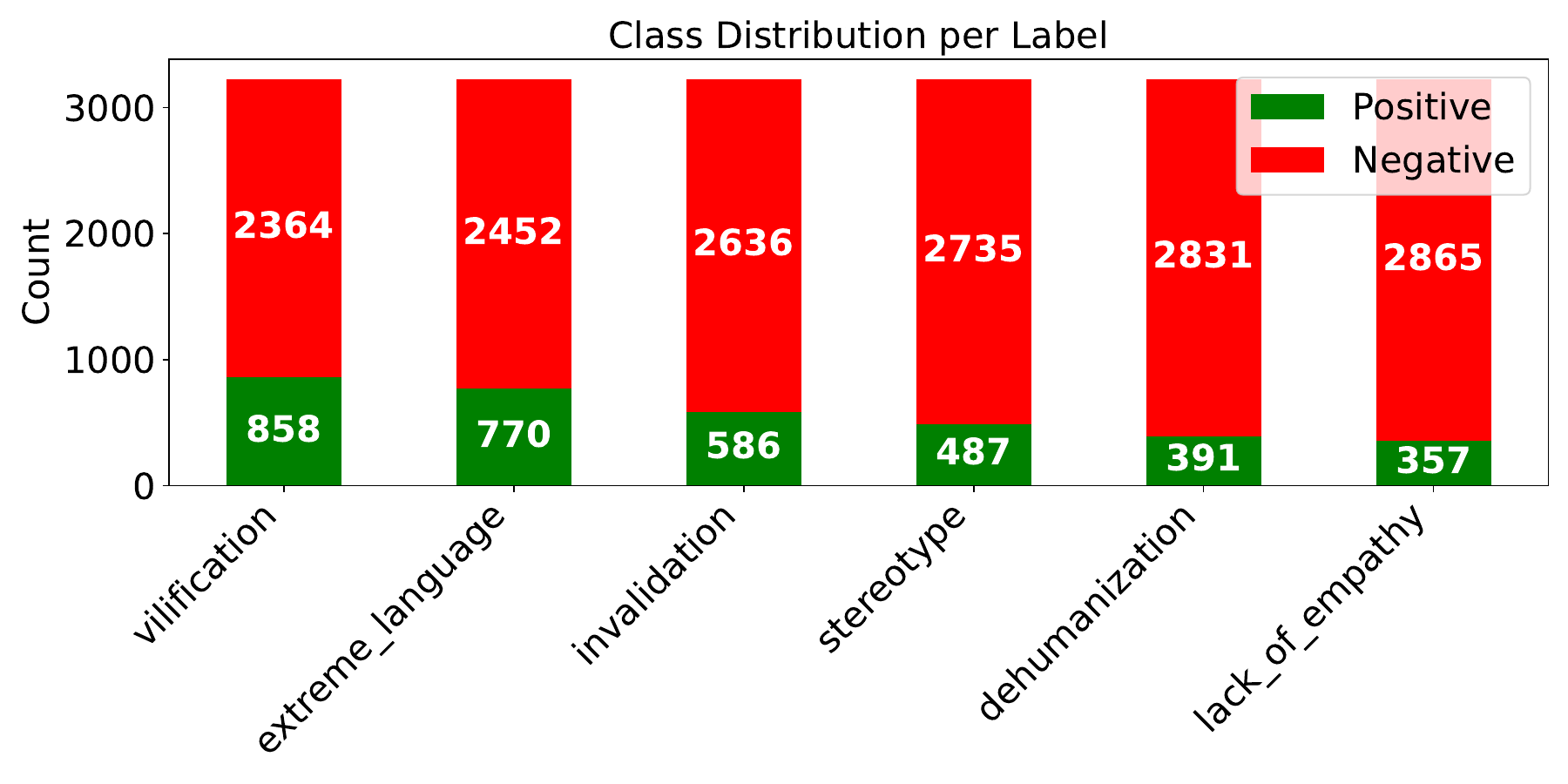}
\caption{Subtask 3 manifestation label distribution.}
\label{fig:manifestation-dist-subtask3}
\end{figure}

\begin{figure}[t]
\centering
\includegraphics[width=0.45\textwidth]{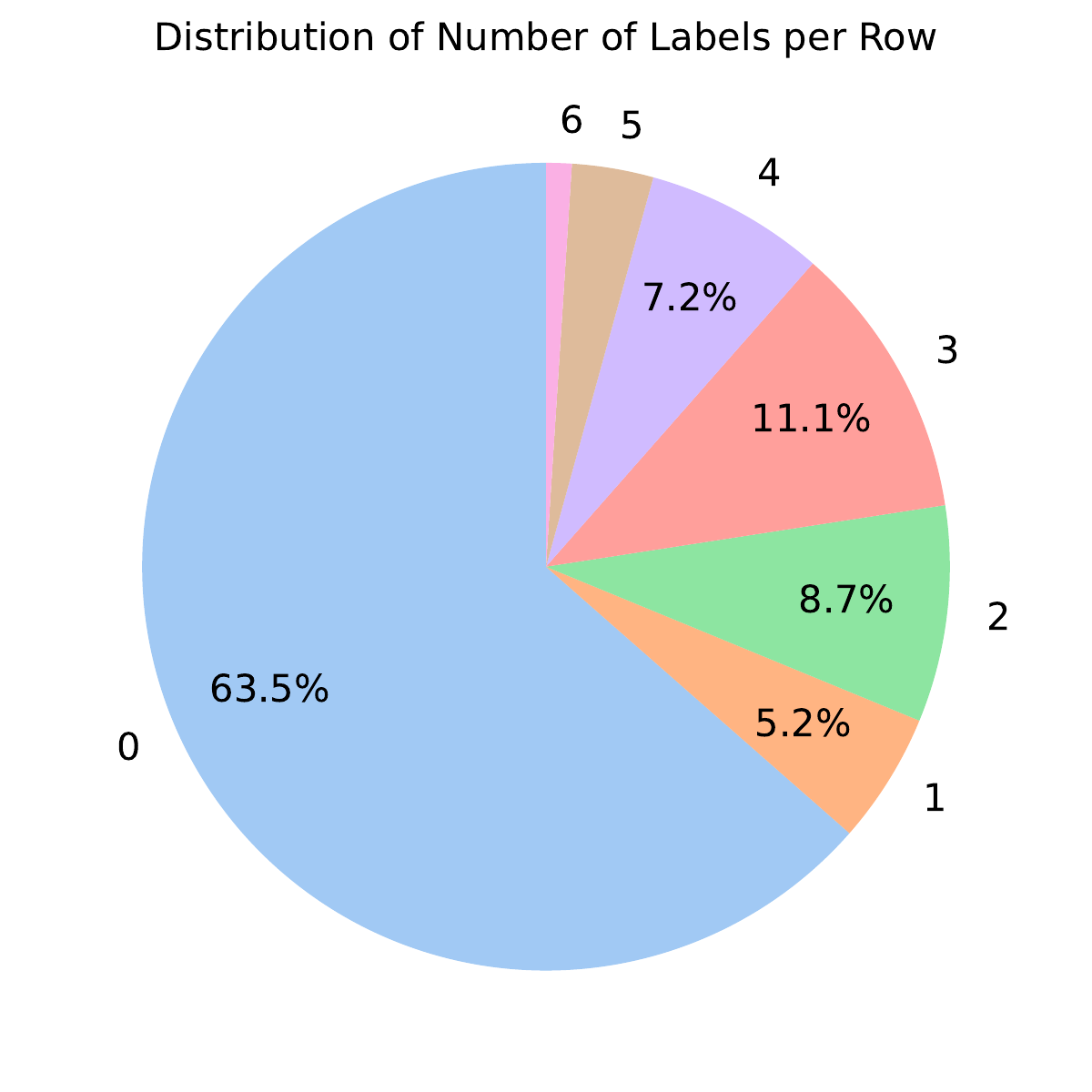}
\caption{Distribution of number of labels per instance in Subtask 3. Most instances (63.5\%) have exactly 0 number of labels, indicating sparse multi-label patterns.}
\label{fig:label-cardinality}
\end{figure}

Figures~\ref{fig:manifestation-dist-subtask3} and~\ref{fig:label-cardinality} show Subtask 3 patterns. Most instances carry exactly 0 labels (63.5\%).

These findings directly motivate the design choices: class-weighted loss to address imbalance, iterative stratified splitting to preserve multi-label distributions \cite{szymanski2017network}, and per-label threshold tuning to optimize macro-F1 across rare categories.

\section{System Overview}

\subsection{Preprocessing}

I apply consistent preprocessing across all subtasks. Emojis are converted to text descriptions using the \texttt{emoji} library's \texttt{demojize} function with space delimiters (e.g., smile emoji → ``smiling face''). URLs, user mentions (@username), and hashtag symbols are removed, though hashtag text content is retained. I lowercase all text and normalize whitespace sequences to single spaces. All texts are truncated or padded to 128 tokens based on length distribution analysis showing 95\% of instances fit within this limit.

\subsection{Model Selection}

I evaluate six Transformer architectures \cite{vaswani2017attention} selected based on three criteria: (1) multilingual capability or Swahili specialization, (2) robustness to noisy social media text, and (3) suitability for moderate-length inputs. The candidates are:

\textbf{Cross-lingual models:}
\begin{itemize}
\item \textbf{TwHIN-BERT} (\texttt{Twitter/twhin-bert-base}) \cite{zhang-etal-2022-twhin}: Pretrained on 7 billion multilingual Twitter tweets with inherent noise handling for informal text, hashtags, and user mentions.
\item \textbf{DistilBERT-multilingual} (\texttt{distilbert-base-multilingual-cased}) \cite{sanh2019distilbert}: A distilled multilingual encoder balancing efficiency and expressiveness across 104 languages.
\item \textbf{mDeBERTa-v3-base} (\texttt{microsoft/mdeberta-v3-base}) \cite{he2021deberta}: Disentangled attention architecture that separately models content and position representations, with strong cross-lingual transfer capabilities.
\end{itemize}

\textbf{African language-focused models:}
\begin{itemize}
\item \textbf{SwahBERT} (\texttt{metabloit/swahBERT}) \cite{martin-etal-2022-swahbert}: Pretrained specifically on Swahili corpora including news, Wikipedia, and social media text.
\item \textbf{AfriBERTa-large} (\texttt{castorini/afriberta\_large}) \cite{ogueji-etal-2021-small}: Pretrained on 11 African languages including Swahili, Hausa, Amharic, and others.
\item \textbf{AfroXLMR-large} (\texttt{Davlan/afro-xlmr-large}) \cite{alabi-etal-2022-adapting}: African language adaptation of XLM-RoBERTa \cite{conneau-etal-2020-unsupervised} with continued pretraining on African language corpora.
\end{itemize}

The rationale for this selection is to test whether language-specific pretraining provides advantages over general multilingual models when downstream task data is limited, and whether social media pretraining (TwHIN-BERT) improves robustness to informal text.

\subsection{Subtask 1: Binary Classification}

I use Hugging Face's\footnote{\url{https://huggingface.co/transformers}} \texttt{AutoModelForSequenceClassification} with \texttt{num\_labels=2}. I perform stratified 80/20 train/validation splits to maintain class balance. Key hyperparameters: learning rate 2e-5, weight decay 0.01, 11 epochs, dropout 0.3 (hidden and attention), batch size 16, gradient accumulation steps 2 (effective batch size 32), 100 warmup steps, maximum gradient norm 1.0, label smoothing 0.1 \cite{szegedy2016rethinking}, and cosine learning rate scheduling. Early stopping with patience 5 halts training if validation macro-F1 does not improve for five consecutive epochs.

Class-weighted loss addresses imbalance by computing per-class weights using scikit-learn's\footnote{\url{https://scikit-learn.org}} \texttt{compute\_class\_weight} with \texttt{balanced} strategy \cite{king-zeng-2001-logistic}:
\[
w_i = \frac{n\_samples}{n\_classes \times n\_samples_i}
\]
where \(w_i\) is the weight for class \(i\), \(n\_samples\) is total training samples, \(n\_classes=2\), and \(n\_samples_i\) is the number of samples in class \(i\). This assigns higher loss contributions to minority classes, encouraging models to learn decision boundaries for underrepresented categories.

\subsection{Subtask 2: Multi-Label Target Types}

Subtask 2 introduces multi-label classification over five target types. I replace stratified splitting with iterative stratified splitting using \texttt{skmultilearn}'s\footnote{\url{https://github.com/scikit-multilearn/scikit-multilearn}} \texttt{iterative\_train\_test\_split} \cite{sechidis2011stratification,szymanski2017network}, which preserves label distribution across multiple labels simultaneously through an iterative algorithm that maintains per-label proportions.

The loss function is \texttt{BCEWithLogitsLoss} with per-label positive class weights:
\[
w_{pos,i} = \frac{n_{neg,i}}{n_{pos,i}}
\]
where \(n_{neg,i}\) and \(n_{pos,i}\) are negative and positive instance counts for label \(i\). This dynamically emphasizes rare labels during training. Configuration: \texttt{MAX\_LENGTH=128}, \texttt{BATCH\_SIZE=32}, \texttt{ACCUM\_STEPS=2}, \texttt{EPOCHS=10}, \texttt{LR=2e-4}, \texttt{WARMUP\_RATIO=0.1}, patience=3.

\subsubsection{Per-Label Threshold Tuning} After training, I apply the threshold optimization procedure described in Algorithm~\ref{alg:threshold-tuning} \cite{saito2015precision}.

\begin{algorithm}
\caption{Per-Label Threshold Tuning}
\label{alg:threshold-tuning}
\begin{algorithmic}[1]
\State \textbf{Input:} Trained model, validation data
\State \textbf{Output:} Per-label thresholds $\{\theta_i\}$

\State \textbf{Stage 1: Coarse Global Search}
\State $\theta^* \gets \arg\max_{\theta \in \{0.20, 0.25, \ldots, 0.80\}} \text{MacroF1}(\theta)$
\State \Comment{Select base threshold maximizing validation macro-F1}

\State
\State \textbf{Stage 2: Fine Per-Label Refinement}
\For{each label $i$}
    \State $\theta_{\min} \gets \max(0.1, \theta^* - 0.15)$
    \State $\theta_{\max} \gets \min(0.9, \theta^* + 0.15)$
    \State $\theta_i \gets \arg\max_{\theta \in [\theta_{\min}, \theta_{\max}]} \text{MacroF1}(\theta \mid \theta_{-i})$
    \State \Comment{Optimize $\theta_i$ in 0.01 increments while fixing other thresholds}
\EndFor

\State \textbf{return} $\{\theta_i\}$
\end{algorithmic}
\end{algorithm}

\subsection{Subtask 3: Multi-Label Manifestations}

Subtask 3 follows the same training loop as Subtask 2 but targets six manifestation labels. I use mDeBERTa-v3-base for both languages based on its superior Subtask 1 performance. Hyperparameters match Subtask 2 except \texttt{LR=2e-5}. Iterative stratified splitting, class-weighted \texttt{BCEWithLogitsLoss}, and two-stage threshold tuning are applied identically.

\subsection{Implementation Details}

All experiments use PyTorch\footnote{\url{https://pytorch.org}} with Hugging Face Transformers. Weights \& Biases\footnote{\url{https://wandb.ai}} (wandb) provides experiment tracking. Random seeds are fixed at 42 for reproducibility. Training was performed on Google Colab (NVIDIA Tesla T4, 15GB VRAM) with runtimes of 20-45 minutes per model.

\section{Experimental Setup}

I use official train set provided by organizers. For Subtask 1, I further partition training sets into 80\% train / 20\% validation using stratified sampling. For Subtasks 2 and 3, iterative stratified splitting ensures validation sets preserve multi-label distributions. All reported validation scores reflect performance on internal validation sets, not official dev sets, to enable fair comparison during development.

Macro-averaged F1 serves as the primary evaluation metric, computing per-label F1 scores and averaging them to weight all labels equally:
\[
\text{Macro-F1} = \frac{1}{|L|} \sum_{i \in L} \text{F1}_i
\]
where \(L\) is the label set and \(\text{F1}_i\) is the F1 score for label \(i\). Micro-averaged F1 is reported supplementally. For multi-label tasks, threshold tuning occurs exclusively on validation data.

Early stopping monitors validation macro-F1 with patience 3--5 epochs depending on subtask. I save checkpoints at each epoch and retain the model with highest validation macro-F1. For Subtask 1, I also experiment with combined English+Swahili training by randomly sampling balanced subsets: (1) sample Swahili rows with label=0 equal to English label=1 count, (2) sample Swahili rows with label=1 equal to English label=0 count, (3) concatenate to achieve 6,444 balanced instances. This tests whether multilingual training improves cross-lingual transfer \cite{pires-etal-2019-multilingual}.

\section{Results}

\subsection{Subtask 1: Binary Detection}

Table~\ref{tab:subtask1} presents validation macro-F1 scores for six models. mDeBERTa-v3-base achieves highest performance (0.8032), followed by AfriBERTa-large (0.7333) and DistilBERT-multilingual (0.7279). Surprisingly, Swahili-specialized models underperform: SwahBERT reaches only 0.6375, and AfroXLMR-large achieves 0.5317.

\begin{table}[t]
\centering
\small
\begin{tabular}{lcc}
\toprule
\textbf{Model} & \textbf{English} & \textbf{Swahili} \\
\midrule
TwHIN-BERT & 0.7279 & 0.7048 \\
DistilBERT-multilingual (Baseline) & 0.7279\textsuperscript{*} & 0.8022\textsuperscript{*} \\
mDeBERTa-v3-base & \textbf{0.8032} & 0.7850 \\
SwahBERT & 0.6375 & 0.8050 \\
AfriBERTa-large & 0.7333 & \textbf{0.8075} \\
AfroXLMR-large & 0.5317 & 0.5788 \\
\bottomrule
\end{tabular}
\caption{Subtask 1 validation macro-F1 scores. \textsuperscript{*}Early stopping at epoch 8. Best scores in bold.}
\label{tab:subtask1}
\end{table}

On official test sets, the best models achieve 0.815 macro-F1 (English) and 0.785 macro-F1 (Swahili), demonstrating strong generalization.

\begin{figure}[t]
\centering
\includegraphics[width=0.45\textwidth]{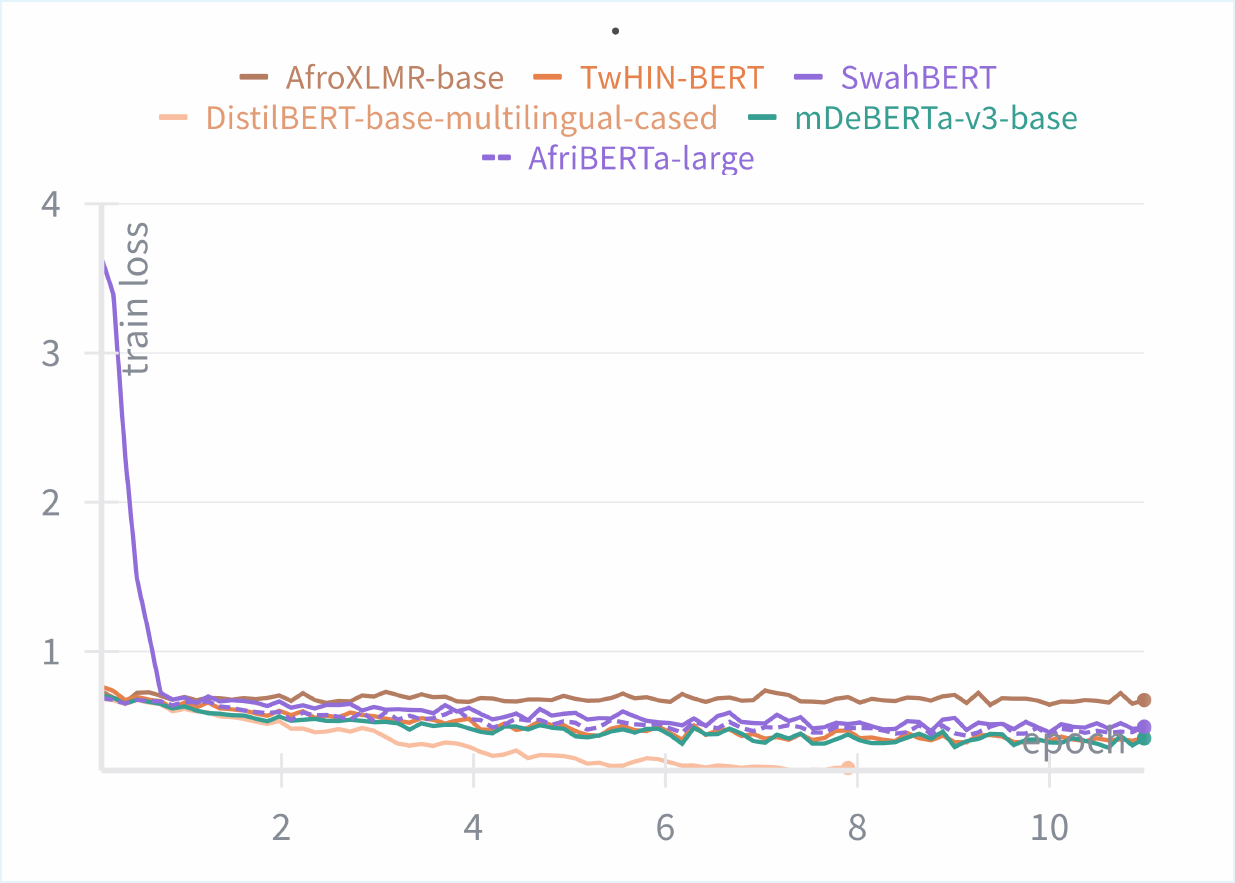}
\caption{Training loss curves for six models on English dataset (Subtask 1). All models show stable convergence.}
\label{fig:subtask1-train-loss-eng}
\end{figure}

\begin{figure}[t]
\centering
\includegraphics[width=0.45\textwidth]{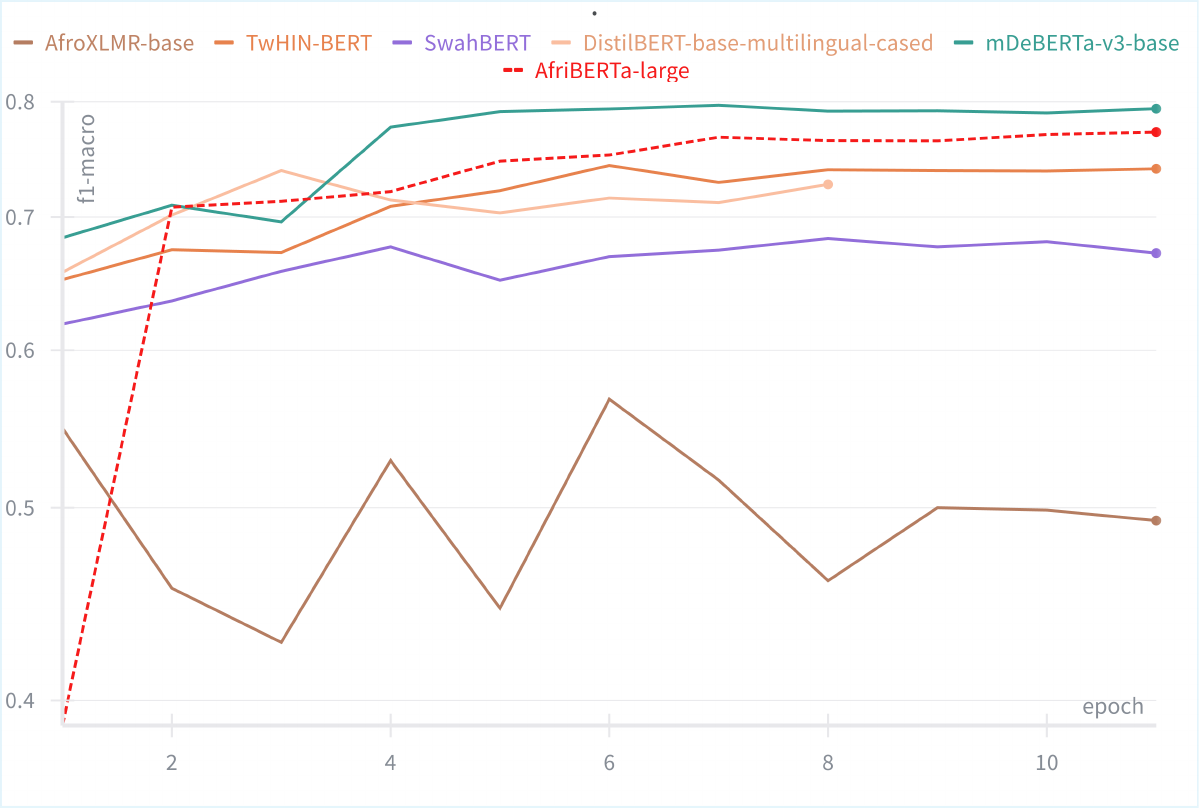}
\caption{Validation F1-macro curves for six models on English dataset (Subtask 1). mDeBERTa-v3-base achieves highest scores with clear separation from alternatives.}
\label{fig:subtask1-val-f1-eng}
\end{figure}

Figure~\ref{fig:subtask1-train-loss-eng} shows training loss curves for all six models on English, with losses converging from 3--4 to 1--2. Figure~\ref{fig:subtask1-val-f1-eng} shows validation F1-macro curves revealing substantial performance stratification (0.50--0.80 range), with mDeBERTa maintaining consistent superiority.

\begin{figure}[t]
\centering
\includegraphics[width=0.45\textwidth]{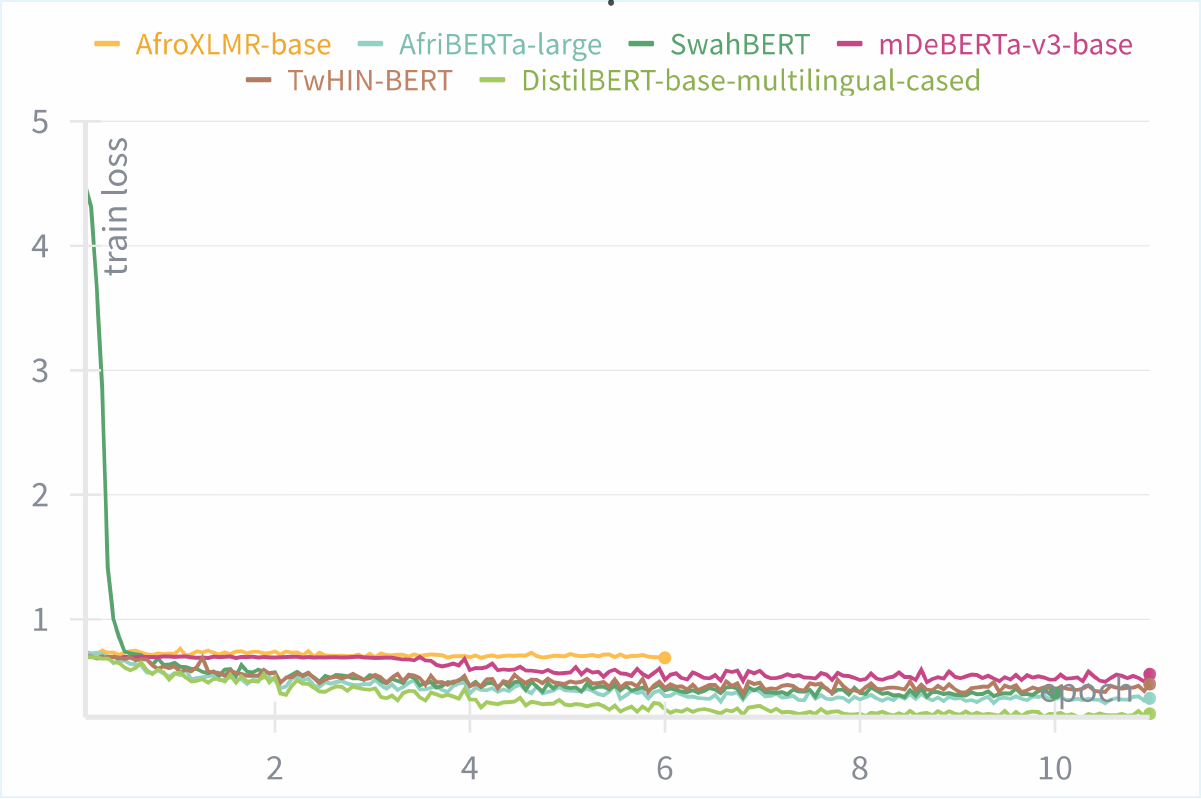}
\caption{Training loss curves for six models on Swahili dataset (Subtask 1). Steeper descent reflects larger corpus providing richer gradient signals.}
\label{fig:subtask1-train-loss-swa}
\end{figure}

\begin{figure}[t]
\centering
\includegraphics[width=0.45\textwidth]{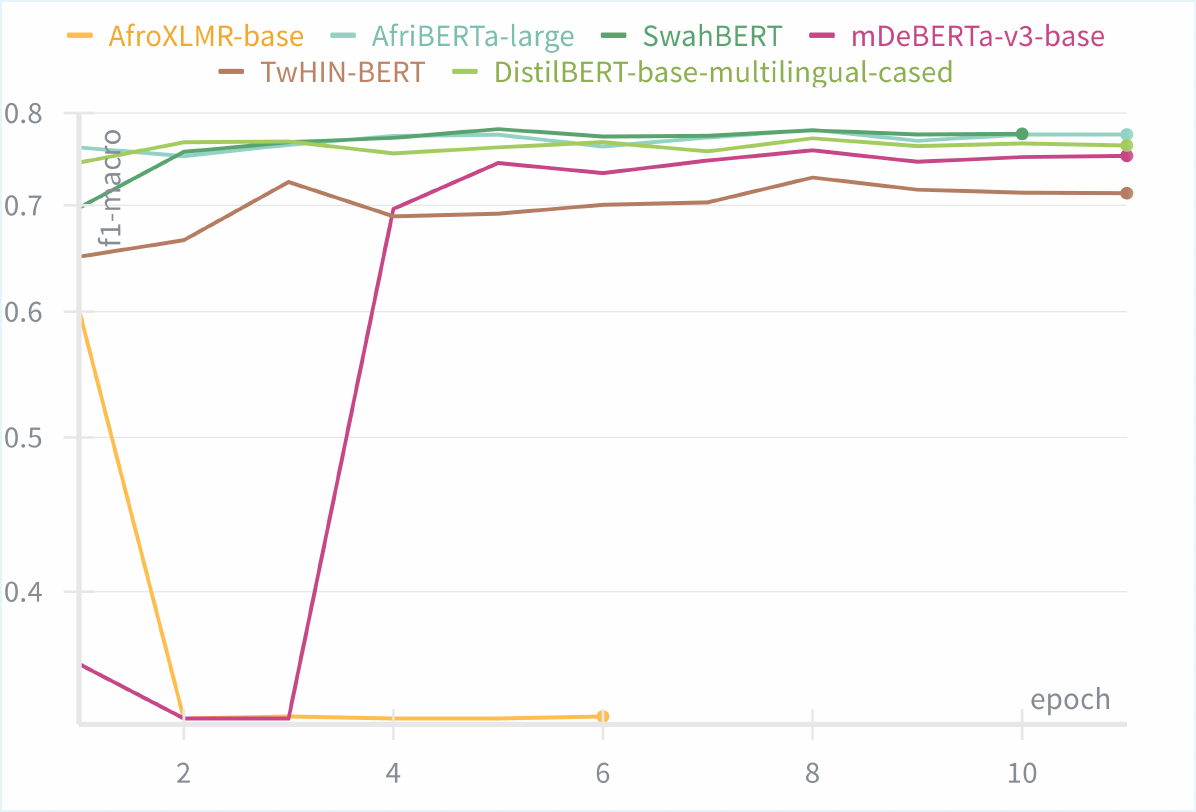}
\caption{Validation F1-macro curves for six models on Swahili dataset (Subtask 1). Smoother trajectories and reduced oscillation compared to English.}
\label{fig:subtask1-val-f1-swa}
\end{figure}

Figures~\ref{fig:subtask1-train-loss-swa} and~\ref{fig:subtask1-val-f1-swa} show Swahili results. Training losses exhibit steeper descent and smoother convergence compared to English due to doubled dataset size. Validation curves show reduced oscillation and clearer performance tiers.

\begin{figure}[t]
\centering
\includegraphics[width=0.45\textwidth]{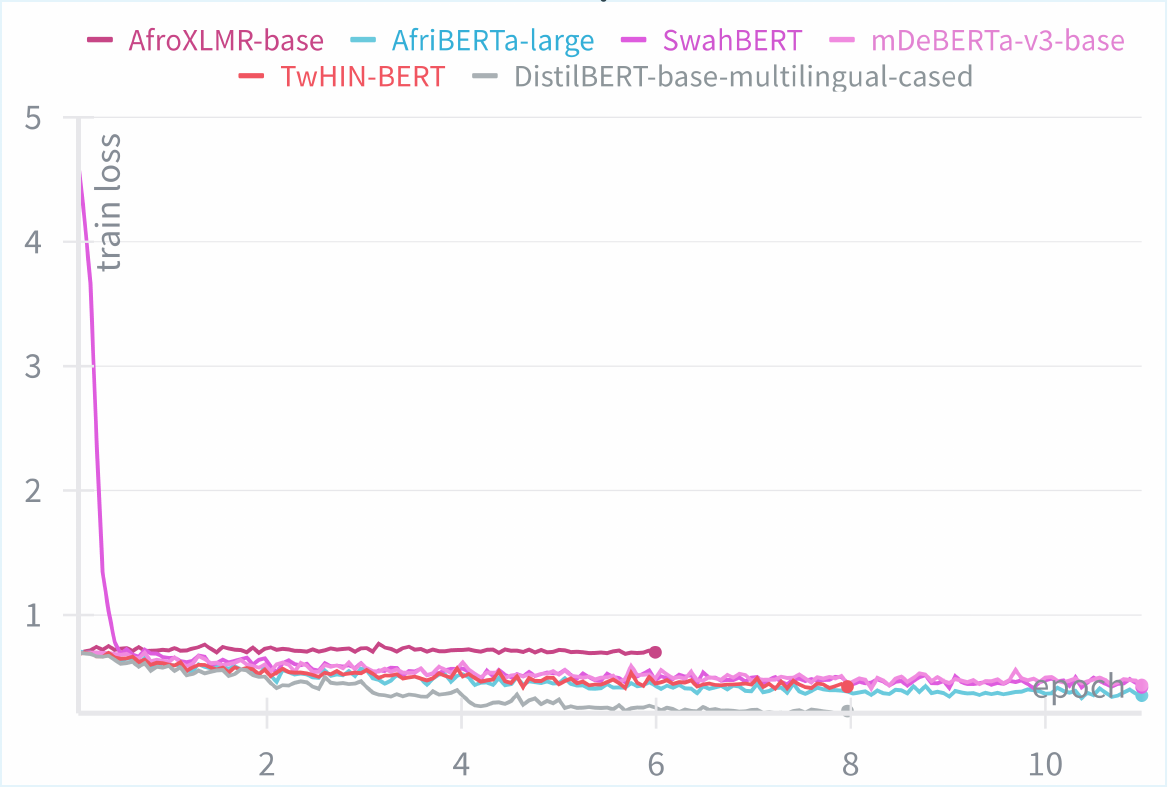}
\caption{Training loss curves for combined English+Swahili dataset (Subtask 1). Comparable convergence to Swahili-only despite increased linguistic diversity.}
\label{fig:subtask1-train-loss-combined}
\end{figure}

\begin{figure}[t]
\centering
\includegraphics[width=0.45\textwidth]{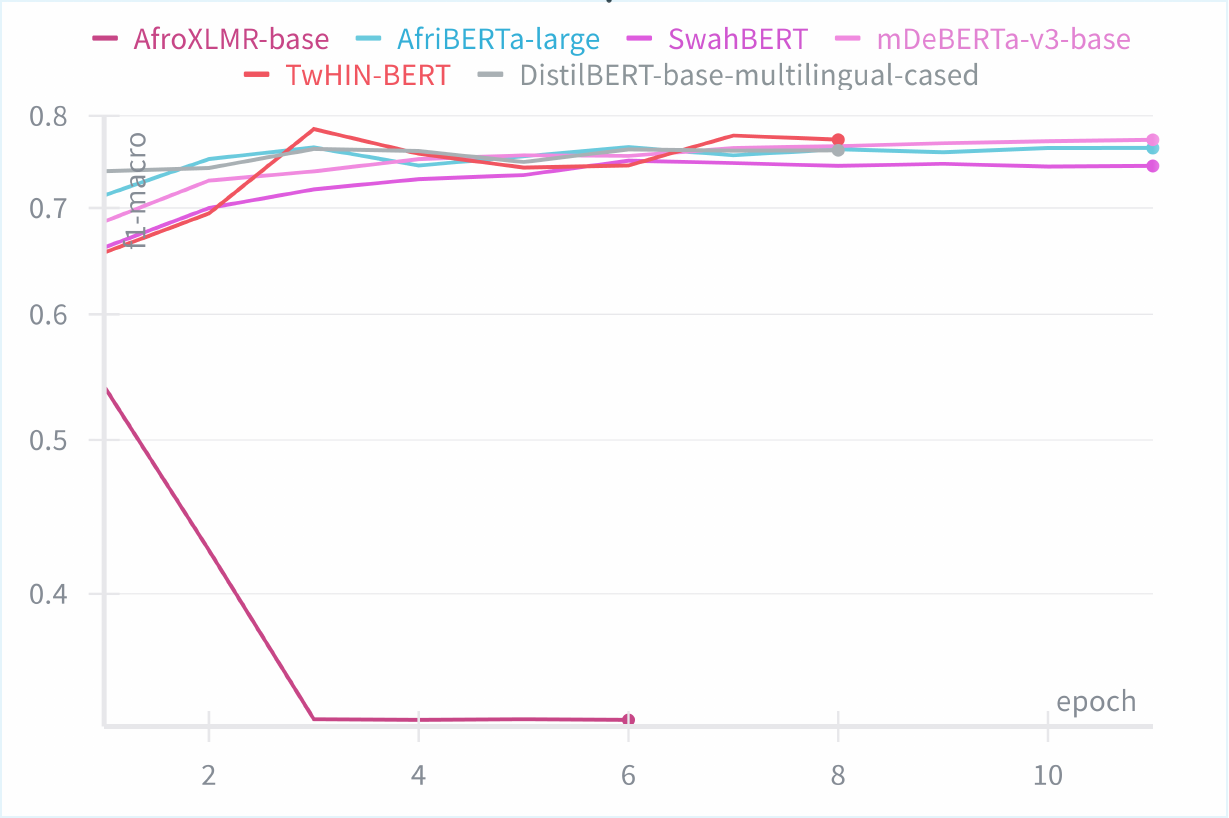}
\caption{Validation F1-macro curves for combined dataset (Subtask 1). Performance degradation indicates negative transfer from naive multilingual training.}
\label{fig:subtask1-val-f1-combined}
\end{figure}

Figures~\ref{fig:subtask1-train-loss-combined} and~\ref{fig:subtask1-val-f1-combined} show combined English+Swahili training results. Validation F1-macro curves reveal performance degradation (5--15 percentage points) relative to single-language experiments, demonstrating negative transfer effects \cite{wang-etal-2020-negative}.

\subsection{Subtask 2: Multi-Label Targets}

Table~\ref{tab:subtask2} shows validation results before and after threshold tuning. AfriBERTa-large on Swahili achieves 0.556 after tuning . SwahBERT on Swahili attains 0.5325 after tuning.

\begin{table}[t]
\centering
\small
\begin{tabular}{lcc}
\toprule
\textbf{Model} & \textbf{Before} & \textbf{After} \\
\midrule
\multicolumn{3}{l}{\textit{English}} \\
DistilBERT-multilingual & 0.245 & \textbf{0.464} \\
AfriBERTa-large & 0.330 & 0.437 \\
\midrule
\multicolumn{3}{l}{\textit{Swahili}} \\
AfriBERTa-large & 0.132 & \textbf{0.556} \\
SwahBERT & 0.210 & 0.325 \\

\bottomrule
\end{tabular}
\caption{Subtask 2 validation macro-F1 before/after threshold tuning.}
\label{tab:subtask2}
\end{table}

\begin{figure}[t]
\centering
\includegraphics[width=0.45\textwidth]{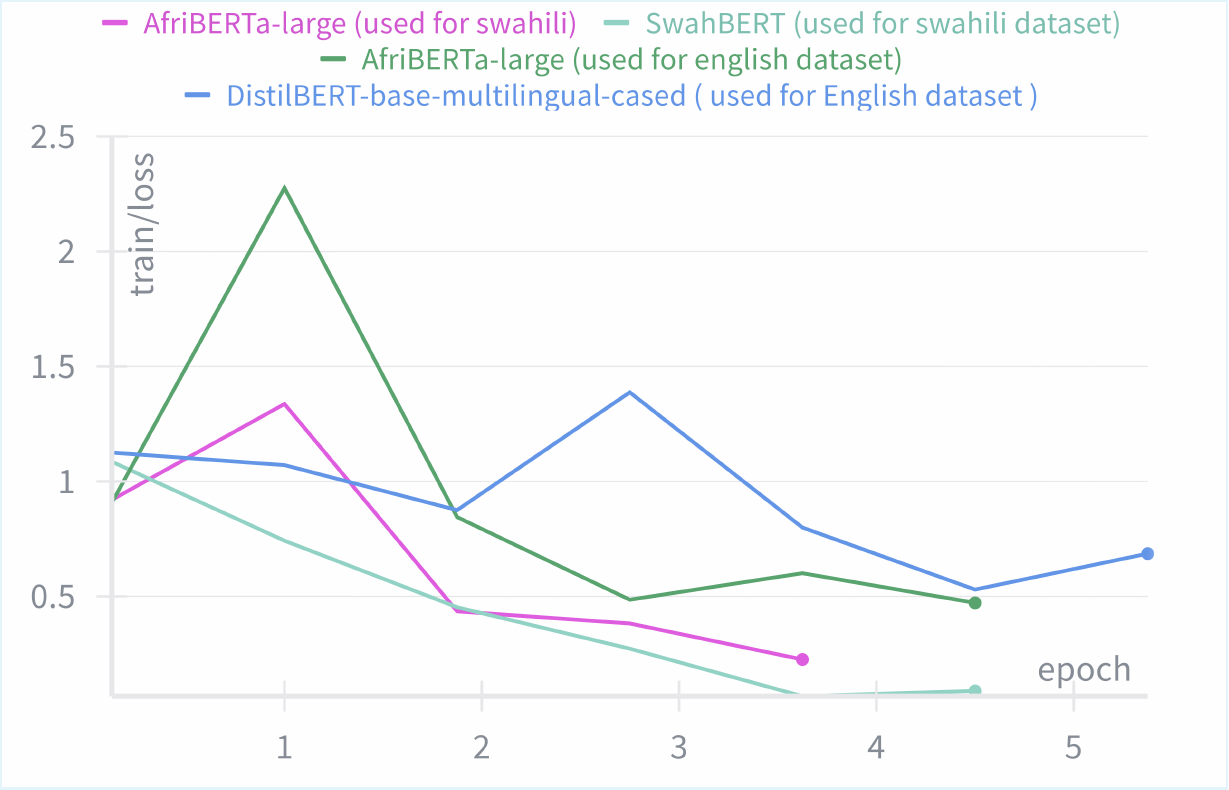}
\caption{Training loss for Subtask 2 configurations. Swahili models achieve lower final losses consistent with larger training sets.}
\label{fig:subtask2-train-loss}
\end{figure}

\begin{figure}[t]
\centering
\includegraphics[width=0.45\textwidth]{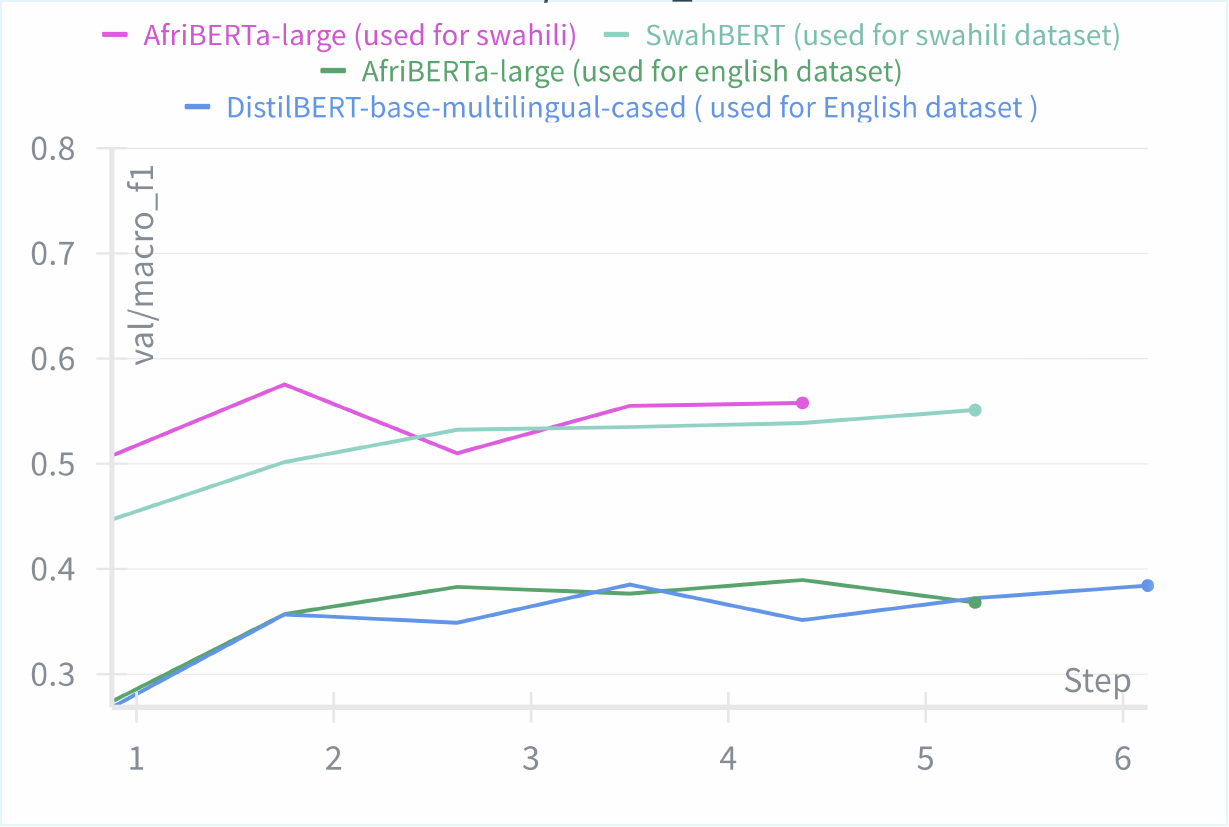}
\caption{Validation F1-macro for Subtask 2. Dramatic 20--25 percentage point gap between Swahili and English models reflects data volume effects.}
\label{fig:subtask2-val-f1}
\end{figure}

Official test performance shows 0.341 macro-F1 (English) and 0.4977 macro-F1 (Swahili), indicating train-test distribution shifts. Figures~\ref{fig:subtask2-train-loss} and~\ref{fig:subtask2-val-f1} show training dynamics.

\begin{figure}[t]
\centering
\includegraphics[width=0.45\textwidth]{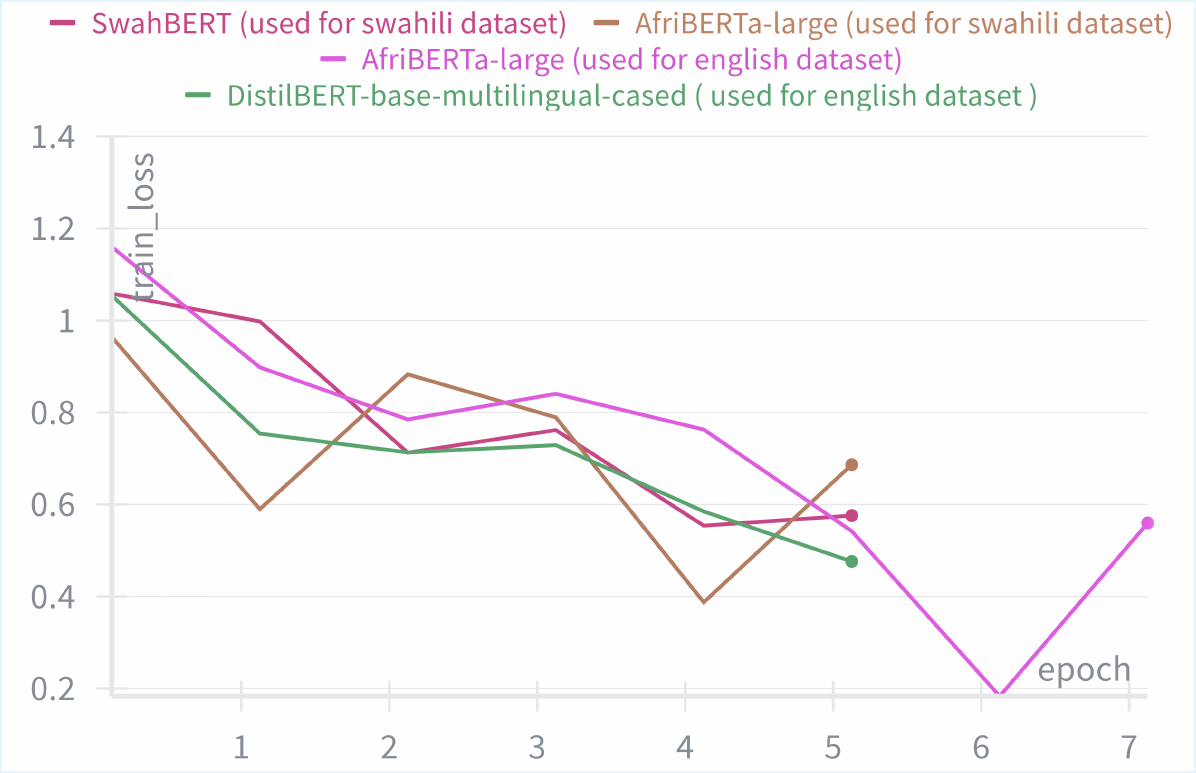}
\caption{Training loss for Subtask 3 configurations. Lower absolute losses (0.2--0.6) than Subtask 2 despite apparent increased task difficulty.}
\label{fig:subtask3-train-loss}
\end{figure}

\begin{figure}[t]
\centering
\includegraphics[width=0.45\textwidth]{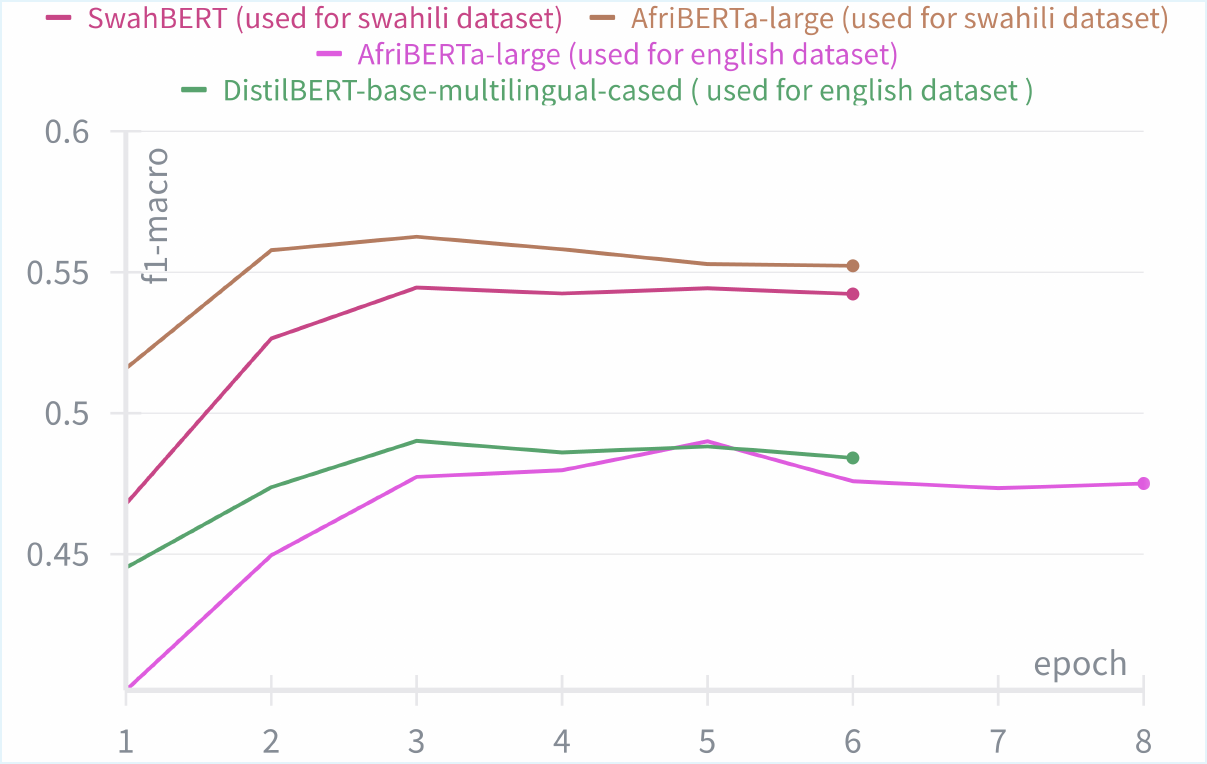}
\caption{Validation F1-macro for Subtask 3. Compressed range (0.45--0.60) represents 15--25 percentage point decline from Subtask 2, indicating fundamental task difficulty.}
\label{fig:subtask3-val-f1}
\end{figure}

\subsection{Subtask 3: Multi-Label Manifestations}

mDeBERTa-v3-base with threshold tuning achieves 0.464 validation macro-F1 on English and comparable scores on Swahili. Official test scores reach 0.464 (English) and 0.556 (Swahili). Figures~\ref{fig:subtask3-train-loss} and~\ref{fig:subtask3-val-f1} show training dynamics.

\subsection{Ablation Studies}

\textbf{Effect of class weighting:} Removing class weights from Subtask 1 degrades English performance from 0.8032 to 0.7214 macro-F1 ($-8.2$ points), as models collapse to majority-class predictions, consistent with findings in imbalanced learning literature \cite{buda2018systematic}.

\textbf{Effect of early stopping:} DistilBERT-multilingual demonstrates the critical role of early stopping: training beyond epoch 8 leads to overfitting with validation F1 declining from 0.7279 to 0.6892 by epoch 11.

\textbf{Effect of threshold tuning:} For Subtask 2, using default 0.5 thresholds yields 0.14 macro-F1, versus 0.556 with tuned thresholds (+26.5 points), validating the two-stage search strategy \cite{saito2015precision}.

\subsection{Error Analysis}

I manually inspect validation errors per subtask to identify systematic failure modes.

\textbf{False Positives (Type I errors).} Most false positives in Subtask 1 occur on heated but non-polarized political discourse. Example: \textit{``This administration has completely failed on every promise they made''} is misclassified as polarized despite expressing policy criticism without out-group vilification. These errors suggest models rely on intensity markers (``completely failed'') rather than semantic understanding of polarization.

\textbf{False Negatives (Type II errors).} False negatives concentrate on implicit polarization and code-switching. Example: \textit{``Those people just don't understand our way of life''} uses the euphemistic dog whistle ``those people'' as an ethnic marker, which models miss without contextual understanding. Code-switched posts mixing English and Swahili (e.g., \textit{``Hawa watu are destroying everything''}) confuse subword tokenizers, which fragment mixed terms into unrecognizable sequences \cite{solorio-etal-2014-overview}.

\textbf{Multi-label confusion.} In Subtask 2, posts addressing political-ethnic intersections are frequently misclassified as purely political. Example: \textit{``Kikuyu politicians are stealing from everyone''} targets both \textit{political} and \textit{racial/ethnic} categories but often receives only \textit{political} label. Per-label F1 analysis confirms rare categories suffer disproportionately: gender/sexual (F1=0.28) and religious (F1=0.31) versus political (F1=0.67).

\textbf{Manifestation overlap.} Subtask 3 struggles to distinguish vilification from extreme language, as both involve intense negative rhetoric. Example: \textit{``These scum should rot in hell''} exhibits both manifestations but models often predict only one due to overlapping lexical markers.

\section{Conclusion}

I present a comprehensive system for multilingual polarization detection, demonstrating the impact of class-weighted loss and per-label threshold tuning on achieving strong performance on binary and multi-label classification. The results reveal the following key findings:

\begin{enumerate}
    \item \textbf{Architectural choice is crucial:} Cross-lingual models outperform Swahili-specialist models by 10--15 percentage points, indicating that model architecture matters more than language-specific pretraining.
    \item \textbf{Threshold tuning boosts multi-label performance:} Per-label threshold tuning provides substantial gains, improving macro-F1 by 20+ points in multi-label tasks under severe class imbalance.
    \item \textbf{Naive multilingual training can cause negative transfer:} Training on multiple languages without careful adaptation degrades performance by 5--15 points compared to single-language models.
\end{enumerate}

Future work should explore the following directions:

\begin{itemize}
    \item \textbf{Data augmentation via back-translation} to expand minority-class coverage \cite{sennrich-etal-2016-improving}.
    \item \textbf{Prompt-based few-shot learning} to leverage large language models' world knowledge \cite{brown2020language}.
    \item \textbf{Contrastive learning objectives} to improve robustness to implicit cues \cite{gao-etal-2021-simcse}.
    \item \textbf{Adapter architectures} to enable positive cross-lingual transfer while preventing negative interference \cite{pfeiffer-etal-2020-mad}.
\end{itemize}

\section*{Limitations}

The system faces several limitations. First, threshold tuning on validation data shows significant overfitting, with test performance dropping 10--15 percentage points below validation scores for multi-label subtasks. Second, I do not address code-switching explicitly despite identifying it as a major error source. Third, the model selection focuses on publicly available pretrained transformers without exploring task-specific pretraining or domain adaptation. Finally, computational constraints limited hyperparameter search scope, and more extensive tuning may yield improvements.

\section*{Acknowledgments}

I thank the SemEval-2025 Polarization Shared Task organizers for developing the datasets and coordinating this shared task evaluation. I acknowledge the use of Google Colab's NVIDIA Tesla T4 GPUs for model training. I am grateful to Dr. Shamsuddeen Muhammad and Dr. Idris Abdulmumin for helpful discussions and feedback on this work. Finally, I thank the Teaching Assistants for their unwavering assistance during the course of this work.

\bibliography{custom}

@inproceedings{vaswani2017attention,
    title = "Attention is All you Need",
    author = "Vaswani, Ashish  and
      Shazeer, Noam  and
      Parmar, Niki  and
      Uszkoreit, Jakob  and
      Jones, Llion  and
      Gomez, Aidan N.  and
      Kaiser, {\L}ukasz  and
      Polosukhin, Illia",
    booktitle = "Advances in Neural Information Processing Systems",
    volume = "30",
    year = "2017",
    url = "https://proceedings.neurips.cc/paper/2017/hash/3f5ee243547dee91fbd053c1c4a845aa-Abstract.html",
}

@inproceedings{he2021deberta,
    title = "{D}e{BERT}a: Decoding-enhanced {BERT} with Disentangled Attention",
    author = "He, Pengcheng  and
      Liu, Xiaodong  and
      Gao, Jianfeng  and
      Chen, Weizhu",
    booktitle = "International Conference on Learning Representations",
    year = "2021",
    url = "https://openreview.net/forum?id=XPZIaotutsD",
}

@inproceedings{sanh2019distilbert,
    title = "{D}istil{BERT}, a distilled version of {BERT}: smaller, faster, cheaper and lighter",
    author = "Sanh, Victor  and
      Debut, Lysandre  and
      Chaumond, Julien  and
      Wolf, Thomas",
    booktitle = "5th Workshop on Energy Efficient Machine Learning and Cognitive Computing - NeurIPS 2019",
    year = "2019",
    url = "https://arxiv.org/abs/1910.01108",
}

@inproceedings{zhang-etal-2022-twhin,
    title = "{T}w{HIN}-{BERT}: A Socially-Enriched Pre-trained Language Model for Multilingual Tweet Representations",
    author = "Zhang, Xinyang  and
      Malkov, Yury  and
      Florez, Omar  and
      Park, Serim  and
      McAuley, Julian  and
      Caverlee, James",
    booktitle = "Proceedings of the 2022 Conference of the North American Chapter of the Association for Computational Linguistics: Human Language Technologies",
    month = jul,
    year = "2022",
    address = "Seattle, United States",
    publisher = "Association for Computational Linguistics",
    url = "https://aclanthology.org/2022.naacl-main.378",
    doi = "10.18653/v1/2022.naacl-main.378",
    pages = "5135--5147",
}

@article{conneau-etal-2020-unsupervised,
    title = "Unsupervised Cross-lingual Representation Learning at Scale",
    author = "Conneau, Alexis  and
      Khandelwal, Kartikay  and
      Goyal, Naman  and
      Chaudhary, Vishrav  and
      Wenzek, Guillaume  and
      Guzm{\'a}n, Francisco  and
      Grave, Edouard  and
      Ott, Myle  and
      Zettlemoyer, Luke  and
      Stoyanov, Veselin",
    journal = "Proceedings of the 58th Annual Meeting of the Association for Computational Linguistics",
    year = "2020",
    pages = "8440--8451",
    url = "https://aclanthology.org/2020.acl-main.747",
    doi = "10.18653/v1/2020.acl-main.747"
}

@inproceedings{martin-etal-2022-swahbert,
    title = "{S}wah{BERT}: Language Model of {S}wahili",
    author = "Mukabana, Martin  and
      Orwa, Daisy  and
      Wambugu, Lawrence  and
      Ogema, Daniel",
    booktitle = "Proceedings of the 2022 Conference of the North American Chapter of the Association for Computational Linguistics: Human Language Technologies",
    month = jul,
    year = "2022",
    address = "Seattle, United States",
    publisher = "Association for Computational Linguistics",
    url = "https://aclanthology.org/2022.naacl-main.23",
    doi = "10.18653/v1/2022.naacl-main.23",
    pages = "303--313"
}

@article{ogueji-etal-2021-small,
    title = "Small Data? No Problem! Exploring the Viability of Pretrained Multilingual Language Models for Low-resourced Languages",
    author = "Ogueji, Kelechi  and
      Zhu, Yuxin  and
      Lin, Jimmy",
    journal = "Proceedings of the 1st Workshop on Multilingual Representation Learning",
    year = "2021",
    pages = "116--126",
    publisher = "Association for Computational Linguistics",
    url = "https://aclanthology.org/2021.mrl-1.11",
    doi = "10.18653/v1/2021.mrl-1.11"
}

@inproceedings{alabi-etal-2022-adapting,
    title = "Adapting Pre-trained Language Models to {A}frican Languages via Multilingual Adaptive Fine-Tuning",
    author = "Alabi, Jesujoba  and
      Amponsah-Kaakyire, Kwabena  and
      Adelani, David  and
      Espa{\~n}a-Bonet, Cristina",
    booktitle = "Proceedings of the 29th International Conference on Computational Linguistics",
    month = oct,
    year = "2022",
    address = "Gyeongju, Republic of Korea",
    publisher = "International Committee on Computational Linguistics",
    url = "https://aclanthology.org/2022.coling-1.382",
    pages = "4336--4349"
}

@inproceedings{fortuna-nunes-2018-survey,
    title = "A Survey on Automatic Detection of Hate Speech in Text",
    author = "Fortuna, Paula  and
      Nunes, S{\'e}rgio",
    booktitle = "ACM Computing Surveys (CSUR)",
    volume = "51",
    pages = "1--30",
    year = "2018",
    publisher = "ACM",
    url = "https://doi.org/10.1145/3232676"
}

@inproceedings{vidgen-derczynski-2020-directions,
    title = "Directions in Abusive Language Training Data, a Systematic Review: Garbage In, Garbage Out",
    author = "Vidgen, Bertie  and
      Derczynski, Leon",
    booktitle = "PLOS ONE",
    volume = "15",
    year = "2020",
    pages = "e0243300",
    url = "https://doi.org/10.1371/journal.pone.0243300"
}

@inproceedings{grimminger-klinger-2021-hate,
    title = "Hate Speech Detection in {G}erman: {W}hat Can We Learn From {E}nglish?",
    author = "Grimminger, Laura  and
      Klinger, Roman",
    booktitle = "Proceedings of the 5th Workshop on Online Abuse and Harms (WOAH 2021)",
    month = aug,
    year = "2021",
    address = "Online",
    publisher = "Association for Computational Linguistics",
    url = "https://aclanthology.org/2021.woah-1.12",
    pages = "101--112"
}

@inproceedings{ousidhoum-etal-2019-multilingual,
    title = "Multilingual and Multi-Aspect Hate Speech Analysis",
    author = "Ousidhoum, Nedjma  and
      Lin, Zizheng  and
      Zhang, Hongming  and
      Song, Yangqiu  and
      Yeung, Dit-Yan",
    booktitle = "Proceedings of the 2019 Conference on Empirical Methods in Natural Language Processing and the 9th International Joint Conference on Natural Language Processing (EMNLP-IJCNLP)",
    month = nov,
    year = "2019",
    address = "Hong Kong, China",
    publisher = "Association for Computational Linguistics",
    url = "https://aclanthology.org/D19-1474",
    doi = "10.18653/v1/D19-1474",
    pages = "4675--4684"
}

@inproceedings{caselli-etal-2021-hatebert,
    title = "{H}ate{BERT}: Retraining {BERT} for Abusive Language Detection in {E}nglish",
    author = "Caselli, Tommaso  and
      Basile, Valerio  and
      Mitrovi{\'c}, Jelena  and
      Granitzer, Michael",
    booktitle = "Proceedings of the 5th Workshop on Online Abuse and Harms (WOAH 2021)",
    month = aug,
    year = "2021",
    address = "Online",
    publisher = "Association for Computational Linguistics",
    url = "https://aclanthology.org/2021.woah-1.3",
    pages = "17--25"
}

@inproceedings{bigoulaeva-etal-2021-cross,
    title = "Cross-Lingual Transfer Learning for Hate Speech Detection",
    author = "Bigoulaeva, Irina  and
      Hangya, Viktor  and
      Fraser, Alexander",
    booktitle = "Proceedings of the First Workshop on Language Technology for Equality, Diversity and Inclusion",
    month = apr,
    year = "2021",
    address = "Kyiv",
    publisher = "Association for Computational Linguistics",
    url = "https://aclanthology.org/2021.ltedi-1.3",
    doi = "10.18653/v1/2021.ltedi-1.3",
    pages = "15--25"
}

@article{chawla2002smote,
    title = "{SMOTE}: Synthetic Minority Over-sampling Technique",
    author = "Chawla, Nitesh V. and Bowyer, Kevin W. and Hall, Lawrence O. and Kegelmeyer, W. Philip",
    journal = "Journal of Artificial Intelligence Research",
    volume = "16",
    pages = "321--357",
    year = "2002",
    url = "https://doi.org/10.1613/jair.953"
}

@inproceedings{lin2017focal,
    title = "Focal Loss for Dense Object Detection",
    author = "Lin, Tsung-Yi and Goyal, Priya and Girshick, Ross and He, Kaiming and Doll{\'a}r, Piotr",
    booktitle = "Proceedings of the IEEE International Conference on Computer Vision",
    pages = "2980--2988",
    year = "2017",
    url = "https://openaccess.thecvf.com/content_iccv_2017/html/Lin_Focal_Loss_for_ICCV_2017_paper.html"
}

@article{buda2018systematic,
    title = "A Systematic Study of the Class Imbalance Problem in Convolutional Neural Networks",
    author = "Buda, Mateusz and Maki, Atsuto and Mazurowski, Maciej A.",
    journal = "Neural Networks",
    volume = "106",
    pages = "249--259",
    year = "2018",
    publisher = "Elsevier",
    url = "https://doi.org/10.1016/j.neunet.2018.07.011"
}

@article{saito2015precision,
  title = {The Precision-Recall Plot Is More Informative than the ROC Plot When Evaluating Binary Classifiers on Imbalanced Datasets},
  author = {Saito, Takaya and Rehmsmeier, Marc},
  journal = {PLOS ONE},
  volume = {10},
  number = {3},
  pages = {e0118432},
  year = {2015},
  url = {https://doi.org/10.1371/journal.pone.0118432}
}

@article{king-zeng-2001-logistic,
    title = "Logistic Regression in Rare Events Data",
    author = "King, Gary and Zeng, Langche",
    journal = "Political Analysis",
    volume = "9",
    number = "2",
    pages = "137--163",
    year = "2001",
    publisher = "Cambridge University Press",
    url = "https://doi.org/10.1093/oxfordjournals.pan.a004868"
}

@inproceedings{sechidis2011stratification,
    title = "On the Stratification of Multi-Label Data",
    author = "Sechidis, Konstantinos and Tsoumakas, Grigorios and Vlahavas, Ioannis",
    booktitle = "Machine Learning and Knowledge Discovery in Databases",
    year = "2011",
    publisher = "Springer",
    pages = "145--158",
    url = "https://doi.org/10.1007/978-3-642-23808-6_10"
}

@inproceedings{szymanski2017network,
    title = "A Network Perspective on Stratification of Multi-Label Data",
    author = "Szyma{\'n}ski, Piotr and Kajdanowicz, Tomasz",
    booktitle = "Proceedings of the First International Workshop on Learning with Imbalanced Domains: Theory and Applications",
    volume = "74",
    pages = "22--35",
    year = "2017",
    url = "http://proceedings.mlr.press/v74/szyma%C5%84ski17a.html"
}

@inproceedings{pires-etal-2019-multilingual,
    title = "How Multilingual is Multilingual {BERT}?",
    author = "Pires, Telmo  and
      Schlinger, Eva  and
      Garrette, Dan",
    booktitle = "Proceedings of the 57th Annual Meeting of the Association for Computational Linguistics",
    month = jul,
    year = "2019",
    address = "Florence, Italy",
    publisher = "Association for Computational Linguistics",
    url = "https://aclanthology.org/P19-1493",
    doi = "10.18653/v1/P19-1493",
    pages = "4996--5001"
}

@inproceedings{wang-etal-2020-negative,
    title = "On Negative Interference in Multilingual Models: Findings and A Meta-Learning Treatment",
    author = "Wang, Zirui  and
      Lipton, Zachary  and
      Tsvetkov, Yulia",
    booktitle = "Proceedings of the 2020 Conference on Empirical Methods in Natural Language Processing (EMNLP)",
    month = nov,
    year = "2020",
    address = "Online",
    publisher = "Association for Computational Linguistics",
    url = "https://aclanthology.org/2020.emnlp-main.359",
    doi = "10.18653/v1/2020.emnlp-main.359",
    pages = "4438--4450"
}

@inproceedings{solorio-etal-2014-overview,
    title = "Overview for the First Shared Task on Language Identification in Code-Switched Data",
    author = "Solorio, Thamar  and
      Blair, Elizabeth  and
      Maharjan, Suraj  and
      Bethard, Steven  and
      Diab, Mona  and
      Ghoneim, Mahmoud  and
      Hawwari, Abdelati  and
      AlGhamdi, Fahad  and
      Hirschberg, Julia  and
      Chang, Alison  and
      Fung, Pascale",
    booktitle = "Proceedings of the First Workshop on Computational Approaches to Code Switching",
    month = oct,
    year = "2014",
    address = "Doha, Qatar",
    publisher = "Association for Computational Linguistics",
    url = "https://aclanthology.org/W14-3907",
    doi = "10.3115/v1/W14-3907",
    pages = "62--72"
}

@inproceedings{szegedy2016rethinking,
    title = "Rethinking the Inception Architecture for Computer Vision",
    author = "Szegedy, Christian and Vanhoucke, Vincent and Ioffe, Sergey and Shlens, Jon and Wojna, Zbigniew",
    booktitle = "Proceedings of the IEEE Conference on Computer Vision and Pattern Recognition",
    pages = "2818--2826",
    year = "2016",
    url = "https://www.cv-foundation.org/openaccess/content_cvpr_2016/html/Szegedy_Rethinking_the_Inception_CVPR_2016_paper.html"
}

@inproceedings{sennrich-etal-2016-improving,
    title = "Improving Neural Machine Translation Models with Monolingual Data",
    author = "Sennrich, Rico  and
      Haddow, Barry  and
      Birch, Alexandra",
    booktitle = "Proceedings of the 54th Annual Meeting of the Association for Computational Linguistics (Volume 1: Long Papers)",
    month = aug,
    year = "2016",
    address = "Berlin, Germany",
    publisher = "Association for Computational Linguistics",
    url = "https://aclanthology.org/P16-1009",
    doi = "10.18653/v1/P16-1009",
    pages = "86--96"
}

@inproceedings{brown2020language,
    title = "Language Models are Few-Shot Learners",
    author = "Brown, Tom B. and Mann, Benjamin and Ryder, Nick and Subbiah, Melanie and Kaplan, Jared and Dhariwal, Prafulla and Neelakantan, Arvind and Shyam, Pranav and Sastry, Girish and Askell, Amanda and others",
    booktitle = "Advances in Neural Information Processing Systems",
    volume = "33",
    pages = "1877--1901",
    year = "2020",
    url = "https://proceedings.neurips.cc/paper/2020/hash/1457c0d6bfcb4967418bfb8ac142f64a-Abstract.html"
}

@inproceedings{gao-etal-2021-simcse,
    title = "{S}im{CSE}: Simple Contrastive Learning of Sentence Embeddings",
    author = "Gao, Tianyu  and
      Yao, Xingcheng  and
      Chen, Danqi",
    booktitle = "Proceedings of the 2021 Conference on Empirical Methods in Natural Language Processing",
    month = nov,
    year = "2021",
    address = "Online and Punta Cana, Dominican Republic",
    publisher = "Association for Computational Linguistics",
    url = "https://aclanthology.org/2021.emnlp-main.552",
    doi = "10.18653/v1/2021.emnlp-main.552",
    pages = "6894--6910"
}

@inproceedings{pfeiffer-etal-2020-mad,
    title = "{MAD-X}: An Adapter-Based Framework for Multi-Task Cross-Lingual Transfer",
    author = "Pfeiffer, Jonas  and
      Vuli{\'c}, Ivan  and
      Gurevych, Iryna  and
      Ruder, Sebastian",
    booktitle = "Proceedings of the 2020 Conference on Empirical Methods in Natural Language Processing (EMNLP)",
    month = nov,
    year = "2020",
    address = "Online",
    publisher = "Association for Computational Linguistics",
    url = "https://aclanthology.org/2020.emnlp-main.617",
    doi = "10.18653/v1/2020.emnlp-main.617",
    pages = "7654--7673"
}
\end{document}